\documentclass[journal]{IEEEtran}
\ifCLASSINFOpdf
\else
   \usepackage[dvips]{graphicx}
\fi
\usepackage{url}

\usepackage{amsmath, amssymb, amsfonts, amsbsy, amsthm, mathtools}
\usepackage{mathrsfs, dsfont, bbm}

\usepackage{nicefrac}
\usepackage{microtype}
\usepackage{soul}
\usepackage{afterpage}
\usepackage{balance}

\usepackage{graphicx}
\usepackage{wrapfig}
\usepackage{booktabs}
\usepackage{adjustbox} 
\usepackage{multirow}
\usepackage{makecell}
\usepackage{xcolor}
\usepackage{nicematrix}
\usepackage{tikz}
\usepackage{tablefootnote}
\usepackage{threeparttable}

\usepackage{algorithm}
\usepackage{algorithmicx}
\usepackage{algpseudocode}

\usepackage[colorlinks,
linkcolor=blue,
anchorcolor=blue,
citecolor=blue]{hyperref}
\usepackage{url}

\theoremstyle{plain}

\theoremstyle{definition}

\theoremstyle{remark}

\def\gap{0.75ex}
\abovedisplayskip\gap
\belowdisplayskip\gap
\abovedisplayshortskip\gap
\belowdisplayshortskip\gap
\title{Optimizing In-Context Learning for Efficient Full Conformal Prediction}

\author{Weicao Deng,~\IEEEmembership{Graduate Student Member,~IEEE,}
        Sangwoo Park,~\IEEEmembership{Member,~IEEE,}\\
        Min Li,~\IEEEmembership{Member,~IEEE,}
        and Osvaldo Simeone,~\IEEEmembership{Fellow,~IEEE}%
\vspace{-2em}
\thanks{W. Deng and M. Li are with the College of Information Science and Electronic Engineering and the Zhejiang Provincial Key Laboratory of Multi-Modal Communication Networks and Intelligent Information, Zhejiang University, Hangzhou, China (e-mail: \{caowd, min.li\}@zju.edu.cn).
S. Park and O. Simeone are with the KCLIP Lab, Centre for Intelligent Information Processing Systems, Department of Engineering, King's College London, London, UK (e-mail: \{sangwoo.park, osvaldo.simeone\}@kcl.ac.uk). 
The work of M. Li was supported in part by National Science and Technology Major Project (Grant No. 2025ZD1302200).
The work of O. Simeone was supported by  an Open Fellowship of the EPSRC (EP/W024101/1) and by the EPSRC project (EP/X011852/1).(\textit{Corresponding author: Min Li}.)}%
}

\begin{document}
\maketitle

\begin{abstract}
Reliable uncertainty quantification is critical for trustworthy AI.
Conformal Prediction (CP) provides prediction sets with distribution-free coverage guarantees, but its two main variants face complementary limitations. Split CP (SCP) suffers from data inefficiency due to dataset partitioning, while full CP (FCP) improves data efficiency at the cost of prohibitive retraining complexity.
Recent approaches based on meta-learning or in-context learning (ICL) partially mitigate these drawbacks. However, they rely on training procedures not specifically tailored to CP, which may yield large prediction sets.
We introduce an efficient FCP framework, termed enhanced ICL-based FCP (E-ICL+FCP), which employs a permutation-invariant Transformer-based ICL model trained with a CP-aware loss.
By simulating the multiple retrained models required by FCP without actual retraining, E-ICL+FCP preserves coverage while markedly reducing both inefficiency and computational overhead.
Experiments on synthetic and real tasks demonstrate that E-ICL+FCP attains superior efficiency-coverage trade-offs compared to existing SCP and FCP baselines.
\end{abstract}


\section{Introduction}
\label{sec:intro}
Reliable quantification of AI decision uncertainty is crucial in modern safety-critical applications \cite{bhatt2021uncertainty,simeone2025conformal}.
Bayesian learning and ensembling can quantify epistemic uncertainty \cite{finn2018probabilistic,yoon2018bayesian,jose2022information}, but their reliability degrades under model or prior misspecification \cite{masegosa2020learning,zecchin2023robust}.
Moreover, exact Bayesian inference is computationally infeasible, while approximations such as Monte Carlo sampling \cite{robert1999monte} and variational inference \cite{blundell2015weight} provide no formal reliability guarantees.

\par \emph{Conformal Prediction} (CP)~\cite{vovk2005algorithmic} is a post-hoc calibration framework that turns any arbitrary predictor -- of a label $y$ given covariates $\mathbf{x}$ -- into a reliable set predictor. 
The set predictor is guaranteed to contain the true label $y$ with probability at least $1 - \alpha$, where $\alpha \in [0,1]$ is a user-specified significance level \cite{angelopoulos2024theoretical}.
To elaborate, assume the availability of a dataset $\mathcal{D}=\{(\mathbf{x}_i,y_i)\}_{i=1}^{n}$ with the $i$-th input $\mathbf{x}_i\in \mathcal{X}$ and the corresponding discrete label $y_i\in\mathcal{Y}=\{1,\dots,K\}$. 
Given a new test input $\mathbf{x}_{n+1}$, the CP predictive set $\mathcal{C}(\mathbf{x}_{n+1})$ is constructed using data $\mathcal{D}$ so as to satisfy the following reliability condition
\begin{equation}
    \text{Pr}(y_{n + 1} \in \mathcal{C}(\mathbf{x}_{n + 1})) \geqslant 1 - \alpha,\label{eq:cp_coverage}
\end{equation}
where the probability is evaluated with respect to the joint distribution of the data. 
The key performance metrics for CP are \emph{coverage} and \emph{inefficiency}: coverage corresponds to the left-hand side of~\eqref{eq:cp_coverage}, while inefficiency denotes the average size of the predictive set, i.e., $\mathbb{E}[|\mathcal{C}(\mathbf{x}_{n+1})|]$.

\par Split CP (SCP) and full CP (FCP) are the two most common CP versions. 
SCP splits the dataset $\mathcal{D}$ into a training set $\mathcal{D}_{\text{tr}}$ and a calibration set $\mathcal{D}_{\text{cal}}$ \cite{vovk2005algorithmic}. 
However, data splitting may lead to an inefficient use of the available data $\mathcal{D}$, resulting in high inefficiency when the available data is limited. 
In contrast, classical FCP uses the entire dataset $\mathcal{D}$ for both training and calibration~\cite{barber2023conformal}, thus achieving lower inefficiency, i.e., smaller predictive sets. 
On the flip side, the complexity of  FCP is significantly higher, requiring retraining of a predictive model for each candidate label $y\in\mathcal{Y}$.

\begin{figure}
    \centering
    \includegraphics[width=1.0\linewidth]{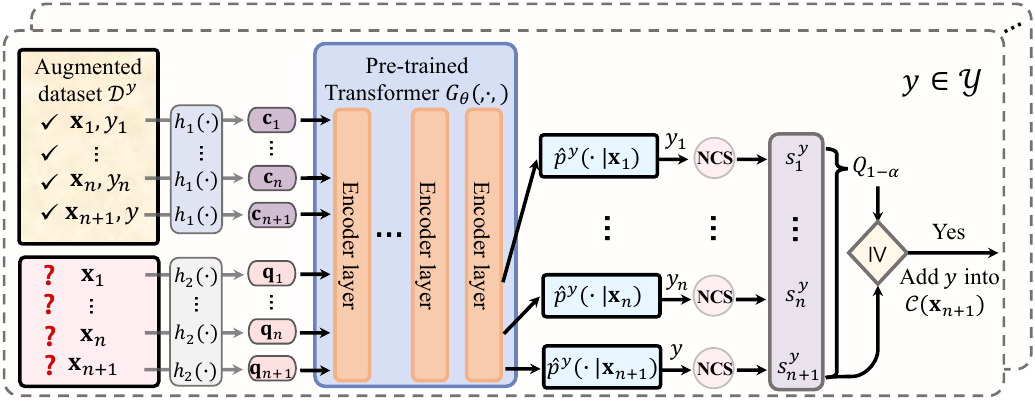}
    \caption{Illustration of the proposed  \emph{enhanced ICL-based FCP} (E-ICL+FCP). E-ICL+FCP employs a permutation-invariant Transformer-based ICL model trained with a CP-aware loss. The model simulates in parallel the retrained models $\{\hat{p}^y(\cdot|\mathbf{x})\}_{y\in \mathcal{Y}}$ required by FCP without implementing any actual retraining as in conventional FCP.}
    \vspace{-15pt}
    \label{fig:ICL+FCP}
\end{figure}

\par Prior work \cite{stutz2021learning} proposed incorporating an average set-size regularizer into the classification loss to mitigate SCP's inefficiency on a single task.
Furthermore, the papers \cite{fisch2021few,park2022pac,park2023few} introduced meta-learning based CP schemes that improve efficiency of CP, while ensuring either an across-task validity guarantee \cite{fisch2021few,park2022pac} or a per-task validity guarantee \cite{park2023few}. 
Proposals to decrease the computational cost of FCP include importance sampling in Bayesian learning \cite{fong2021conformal} and in-context learning (ICL) \cite{huang2025predictions}.

\par ICL is a recent meta-learning paradigm \cite{chen2023learning} that builds upon pre-trained sequence models, like Transformers \cite{fang2025rethinking}, enabling per-task adaptation without any parameter update. 
In this context, recent works \cite{huang2025predictions,mukherjeepredictive} were the first to employ ICL models to implement FCP \cite{huang2025predictions}, as well as the related cross-validation CP \cite{mukherjeepredictive}. 
However, these works rely on conventional ICL training that is not optimized for CP, which may lead to larger (more conservative) prediction sets than needed.

\par In this work, we address this shortcoming of existing ICL-based CP methods by introducing a meta-training methodology explicitly tailored for FCP, aiming to yield more efficient predictive sets while provably meeting the coverage condition \eqref{eq:cp_coverage}.
The proposed scheme, referred to as \emph{enhanced ICL-based FCP} (E-ICL+FCP), meta-trains a permutation-invariant ICL model to minimize the weighted sum of the prediction loss and of a differentiable approximation of the prediction set size \cite{park2023few}. 
Numerical results demonstrate that E-ICL+FCP achieves the desired coverage \eqref{eq:cp_coverage} with smaller predictive sets than ICL-based CP methods; exhibiting higher computational efficiency as compared to conventional, non-ICL-based,  SCP and FCP baselines.

\vspace{-0.2cm}

\section{Background}
\label{sec:Background}
This section presents background on SCP, FCP, and ICL.

\subsection{Split Conformal Prediction}
SCP splits the available data $\mathcal{D}=\{(\mathbf{x}_i,y_i)\}_{i=1}^n$ into a training set $\mathcal{D}_{\text{tr}}= \{(\mathbf{x}_i,y_i)\}_{i=1}^l$ and a calibration set $\mathcal{D}_{\text{cal}}= \{\mathbf{x}_i,y_i\}_{i=l+1}^n$, where $l+m=n$. 
The training dataset $\mathcal{D}_{\text{tr}}$ is used to train a predictor $\hat{p}(y|\mathbf{x})$ of the label $y$ given the covariates $\mathbf{x}$. 
The predictor $\hat{p}(y|\mathbf{x})$ is used to define a \emph{non-conformity score} (NCS) function $s(\mathbf{x}, y)$ that quantifies how ``unusual'' a sample $(\mathbf{x}, y)$ is compared to the training data. 
A common NCS is the log-loss $s(\mathbf{x},y)=-\log(\hat{p}(y|\mathbf{x}))$.
More generally, the NCS can be a function of $y$ and the entire distribution $\hat{p}(\cdot|\mathbf{x})$ \cite{angelopoulos2024theoretical}, but, to simplify the presentation, hereafter we will adopt the log-loss.

\par SCP computes the NCS values on the calibration set, obtaining the set $\mathcal{S} = \{\,s(\mathbf{x},y) : (\mathbf{x},y) \in \mathcal{D}_{\text{cal}}\,\} \cup \{\infty\}$, with an added fictitious data point having an infinite  NCS. 
The prediction set satisfying the coverage condition \eqref{eq:cp_coverage} is then obtained as 
\begin{equation}
    \mathcal{C}(\mathbf{x}_{n+1}) = \{\,y \in \mathcal{Y}: s(\mathbf{x}_{n+1}, y) \leqslant Q_{1-\alpha}(\mathcal{S})\},
\end{equation}
where $Q_{1-\alpha}(\cdot)$ returns the $(1-\alpha)$-quantile of the set $\mathcal{S}$. 
SCP is data-inefficient because of the need to partition data into disjoint training and calibration data sets.

\subsection{Full Conformal Prediction}
\label{subsubsec:full_cp}
In FCP, for each possible label $y \in \mathcal{Y}$, we form an augmented dataset $\mathcal{D}^{y} = \mathcal{D} \cup \{(\mathbf{x}_{n+1}, y)\}$. 
Then, a model $\hat{p}^{y}(\cdot|\mathbf{x})$ is trained on each augmented dataset $\mathcal{D}^y$. 
The training algorithm is assumed to be permutation-invariant with respect to the data in $\mathcal{D}^y$, meaning that it yields the same model regardless of the order of the samples in $\mathcal{D}^y$. 
This condition is  satisfied by standard stochastic gradient-based training algorithms that treat all training samples in the same way \cite{amari1993backpropagation}.

\par Using the $|\mathcal{Y}|$ trained models $\{\hat{p}^y(\cdot|\mathbf{x})\}_{y\in\mathcal{Y}}$, FCP computes the NCS values as $s_i^y = -\log(\hat{p}^y(y_i|\mathbf{x}_i))$ for $i=1,\dots,n$ 
and $s_{n+1}^y = -\log(\hat{p}^y(y|\mathbf{x}_{n+1}))$ for each $y \in \mathcal{Y}$. 
The predictive set for $\mathbf{x}_{n+1}$ is then obtained by including all labels $y$ whose NCS $s_{n+1}^y$ is no greater than the $(1-\alpha)$-quantile of the set $\{s_i^y\}_{i=1}^{n+1}$:
\begin{equation}
    \mathcal{C}(\mathbf{x}_{n+1}) = \left\{ y \in \mathcal{Y}: s_{n+1}^y \leqslant Q_{1-\alpha}\left(\{ s_i^y \}_{i=1}^{n+1}\right) \right\}.
    \label{eq:full_prediction_set}
\end{equation}
FCP generally requires retaining for each candidate label $y$, requiring compute- and memory-intensive gradient descent steps at inference time.

\subsection{In-Context Learning}
\label{subsubsec:icl}
Unlike conventional meta-learning methods \cite{chen2023learning}, ICL exploits latent patterns in the context samples to adapt to new tasks without the need for explicit fine-tuning of model parameters for each task \cite{min2021metaicl}. 
Given $n$ context examples $\mathcal{D}_{\text{con}} = \{(\mathbf{x}_i, y_i)\}_{i=1}^{n}$ and a query input $\mathbf{x}_{n+1}$, an ICL model, $G_{\boldsymbol{\theta}}$, parameterized by vector  $\boldsymbol{\theta}$ produces a predictive  probability vector for $\mathbf{x}_{n+1}$ as the output
\begin{equation}
{\hat{p}(\cdot|\mathbf{x}_{n+1})=G_{\boldsymbol{\theta}}\left(\mathcal{D}_\text{con},\mathbf{x}_{n+1}\right)}.\label{eq:conventional_ICL_model}
\end{equation} The distribution $\hat{p}(\cdot|\mathbf{x}_{n+1})$ approximates the  posterior distribution of the label $y_{n+1}$ given $\mathcal{D}_\text{con}$ and $\mathbf{x}_{n+1}$ \cite{xie2021explanation}.

Conventionally, the ICL model $G_{\boldsymbol{\theta}}$ is pre-trained using data from $T$ auxiliary tasks \cite{wies2023learnability}. 
For each task $t=1,\dots,T$, the context set $\mathcal{D}_{\text{con}}^{t}=\{\mathbf{x}_i^t, y_i^t\}_{i=1}^n$ and the query pair $(\mathbf{x}_{n+1}^t, y_{n+1}^t)$ are drawn i.i.d. from a task-specific distribution. 
The ICL model parameters $\boldsymbol{\theta}$ are optimized by minimizing a loss function  $\ell(\hat{p}(\cdot|\mathbf{x}_{n+1}^t),y_{n+1}^t)$, such as the log-loss $ -\log (\hat{p}(y_{n+1}^t|\mathbf{x}_{n+1}^t))$, averaged over all $T$ tasks. 
This yields the meta-training objective
\begin{equation}
L(\boldsymbol{\theta})= \frac{1}{T}\sum\nolimits_{t=1}^{T} \ell(G_{\boldsymbol{\theta}}(\mathcal{D}_{\text{con}}^{t},\mathbf{x}_{n+1}^{t}), y_{n+1}^{t}).\label{eq:classical pretrained loss}
\end{equation}

\section{Optimizing ICL for Efficient Full Conformal Prediction}
\label{sec:optimized_icl}
Classical FCP suffers from high computational complexity because it requires retraining a separate model $\hat{p}^{y}(\cdot | \mathbf{x})$ for each candidate label $y \in \mathcal{Y}$ of a given test input $\mathbf{x}_{n+1}$. 
In contrast, as discussed in Sec.~\ref{subsubsec:icl}, an ICL model can produce a predictive distribution \eqref{eq:conventional_ICL_model} for a new input using a set of demonstration samples  without needing  additional training. 
Based on this property, building on prior works \cite{huang2025predictions,mukherjeepredictive}, in this section we introduce a novel framework for the implementation and design of ICL-based FCP using a tailored meta-learning strategy.

\subsection{ICL-based FCP}
\label{subsubsec:icl_fcp}
As illustrated in Fig.~\ref{fig:ICL+FCP}, the proposed E-ICL+FCP runs $|\mathcal{Y}|$ copies of the same Transformer model $G_{\boldsymbol{\theta}}$ in parallel to produce the predictive distributions $\hat{p}^{y}(\cdot|\mathbf{x})$ based on each data set $\mathcal{D}^y$ for all $y\in\mathcal{Y}$. These models are used as explained in Sec. \ref{subsubsec:full_cp} to evaluate the NCSs   $\{s_i^y\}_{i=1}^{n+1}$ and $s^y_{n+1}$ and compute the FCP set predictor \eqref{eq:full_prediction_set}.

\par To compute the predictive distribution $\hat{p}^{y}(\cdot|\mathbf{x})$, the $n+1$ samples in the augmented dataset $\mathcal{D}^y$ are each encoded as context tokens $\{\mathbf{c}_1,\dots,\mathbf{c}_n,\mathbf{c}_{n+1}\}$ via an embedding  model $h_1(\cdot)$, while the corresponding $n+1$ queries are encoded as query tokens $\{\mathbf{q}_1,\dots,\mathbf{q}_n,\mathbf{q}_{n+1}\}$ via another model $h_2(\cdot)$. 
These tokens are concatenated as $[\mathbf{c}_1,\dots,\mathbf{c}_n,\mathbf{c}_{n+1},\mathbf{q}_1,\dots,\mathbf{q}_n,\mathbf{q}_{n+1}]$ and fed into the Transformer model $G_{\boldsymbol{\theta}}$, which is composed of a number of layers.

\par To satisfy the exchangeability assumption of FCP, which requires that any permutation of samples in the dataset $\mathcal{D}^y$ leaves the NCSs unchanged, the Transformer's query outputs must be permutation-invariant with respect to the ordering of the context tokens.
To this end, the $(2n+2)\times (2n+2)$ attention mask is given by \cite{raffel2020exploring,fang2025rethinking}
\begin{equation}
\small
\mathbf{M} =
\begin{bNiceMatrix}[first-row,first-col]
    & \text{context} & \text{query} \\
\text{context} 
    & \Block{1-1}{\mathbf{0}_{(n+1)\times(n+1)}} 
    & \Block{1-1}{-\infty \cdot \mathbf{1}_{(n+1)\times(n+1)}} \\
\text{query}   
    & \Block{1-1}{\mathbf{0}_{(n+1)\times(n+1)}} 
    & \Block{1-1}{-\infty(\mathbf{1}_{(n+1)\times(n+1)}-\mathbf{I}_{n+1})}
\end{bNiceMatrix},\label{eq:M}
\end{equation}
where $\mathbf{1}_{a\times b}$ represents the all-one matrix of size $a\times b$, while $\mathbf{I}_a$ is the $a\times a $ identity matrix. 
We use the convention $\infty \cdot 0 = 0$. This mask ensures that each query token can attend only to the context tokens and to itself, while all context tokens can attend to one another, ensuring permutation invariance with respect to context tokens \cite[Proposition~3.4]{fang2025rethinking}.

\par The last $n+1$ outputs of the Transformer model correspond to the $n+1$ predictive distribution $\{\hat{p}^y(\cdot|\mathbf{x}_i)\}_{i=1}^{n+1}$. 
Based on $\{\hat{p}^y(\cdot|\mathbf{x}_i)\}_{i=1}^{n+1}$, the NCS computation and prediction set construction follow Sec.~\ref{subsubsec:full_cp} as explained above.

\subsection{Enhanced ICL-Based FCP}
\label{sec:optimizing_the_icl_model_for_fcp}
Conventional ICL models are not designed for computing NCS values, and thus they often produce overly large predictive sets. 
Inspired by \cite{stutz2021learning}, which addressed SCP, we aim to design a pre-training procedure that directly minimizes the expected size of the FCP predictive set \eqref{eq:full_prediction_set}.

\subsubsection{Meta-Training Setting}
As discussed in Sec.~\ref{subsubsec:icl}, assume the availability of data from $T$ auxiliary tasks. 
The goal is to optimize the parameter vector $\boldsymbol{\theta}$ of the Transformer model $G_{\boldsymbol{\theta}}$ so as to minimize the average predictive set size. 
Hereafter, we denote the predictive set generated by ICL-based FCP for task $t$ as $\mathcal{C}_{\boldsymbol{\theta}}(\mathbf{x}^t_{n+1})$, emphasizing the dependence on parameters $\boldsymbol{\theta}$. 
According to \eqref{eq:full_prediction_set}, the predictive set size can be expressed as
\begin{equation}
|\mathcal{C}_{\boldsymbol{\theta}}(\mathbf{x}^t_{n+1})|=\sum\nolimits_{y\in\mathcal{Y}}\mathds{1}(s_{n+1}^{t,y}\leqslant Q_{1-\alpha}(\{s_i^{t,y}\}_{i=1}^{n+1})),\label{eq:prediction_set_size_ICL}
\end{equation}where $\mathds{1}(\cdot)$ is the indicator function and the NCSs for task $t$ are denoted as $\{s_{i}^{t,y}\}$ and $s_{n+1}^{t,y}$.  
Therefore, one could in principle address the meta-learning problem
\begin{equation}
    \min_{\boldsymbol{\theta}}\sum\nolimits_{t=1}^{T}|\mathcal{C}_{\boldsymbol{\theta}}(\mathbf{x}_{n+1}^{t})|\label{eq:min_ineff}
\end{equation}
of minimizing the average predictive set size across all tasks. 
However, this is challenging due to the non-differentiability of the objective \eqref{eq:prediction_set_size_ICL}. 
To address this issue, in a manner similar to \cite{stutz2021learning}, we introduce a smooth approximation of the objective \eqref{eq:min_ineff} in the rest of this section.

\subsubsection{Soft Quantile, Soft Indicator, and Differentiable Meta-Training Function}
The soft $(1-\alpha)$-quantile of the set $z_1,\dots,z_n$ is defined as \cite{park2023few}
\begin{equation}
\hat{Q}_{1-\alpha}\left(\{z_i\}_{i=1}^{n}\right)=\sum_{j=1}^{n}z_j\frac{e^{-\rho_{1-\alpha}(z_j;\{z_j\}_{j=1}^{n})/c_q}}{\sum_{i=1}^{n}e^{-\rho_{1-\alpha}(z_i;\{z_j\}_{j=1}^{n})/c_q}}, \label{eq:soft_quantile}
\end{equation}
where parameter $c_q>0$ controls the smoothness of function \eqref{eq:soft_quantile}, with a larger $c_q$ yielding a larger smoothness level. 
The function $\rho_{1-\alpha}(z;\{z_j\}_{j=1}^{n})$ in \eqref{eq:soft_quantile} is the \emph{pinball loss}, given by 
\begin{equation}
\rho_{1-\alpha}(z;\{z_j\}_{j=1}^{n}) = \alpha \sum_{j=1}^{n} [\,z-z_j\,]_+ + (1-\alpha) \sum_{j=1}^{n} [\,z_j-z\,]_+,\notag
\end{equation}
with $[x]_+=\max(x,0)$.

\par Furthermore, given a threshold condition $d \leqslant \tau$, the soft indicator function $\sigma(d, \tau)$ is defined as $ {1}/{(1 + e^{-(d - \tau)/\kappa})}$, where $\kappa>0$ controls the sharpness of the transition. 
As $\kappa \to 0$, the function $\sigma(d, \tau)$ approaches the hard indicator function $\mathds{1}(d \leqslant \tau)$.

\par Using the soft $(1-\alpha)$-quantile and soft indicator, a smooth surrogate for the meta-training  loss  \eqref{eq:min_ineff} is obtained as
\begin{equation}
    L_{\text{ineff}}(\boldsymbol{\theta})=\sum\nolimits_{t=1}^{T}\sum\nolimits_{y\in\mathcal{Y}}\sigma(s_{n+1}^{t,y}, \hat{Q}_{1-\alpha}(\{s_{i}^{t,y}\}_{i=1}^{n+1})).\label{eq:surrogate loss}\notag
\end{equation}

\par To promote inclusion of the true label $y_{n+1}^t$ in the predictive set, driving the corresponding value of the soft indicator towards~1, we incorporate in the meta-training criterion also the classification-related term \cite{stutz2021learning}
\begin{equation}
L_{\text{class}}(\boldsymbol{\theta})
= \sum\nolimits_{t=1}^{T} \Big(1-\sigma\!\left(s_{n+1}^{t,y_{n+1}^t}, \hat{Q}_{1-\alpha}(\{s_{i}^{t,y}\}_{i=1}^{n+1})\right)\Big). \label{eq:class_loss}\notag
\end{equation}

Overall, the meta-training objective for the Transformer model $G_{\boldsymbol{\theta}}$ is given by the \emph{CP-aware loss}, which is defined as
\begin{equation}
L(\boldsymbol{\theta}) = L_{\text{ineff}}(\boldsymbol{\theta})+\lambda L_{\text{class}}(\boldsymbol{\theta}), \label{eq:loss_combined}
\end{equation}
where $\lambda>0$ is a hyperparameter balancing the two loss terms.

\section{Experiments}
\label{sec:experiments}
In this section, we evaluate the proposed E-ICL+FCP against ICL+FCP \cite{huang2025predictions} and a range of baselines spanning joint learning, meta-learning, and ICL, under both SCP and FCP. 
Results are reported on a synthetic task \cite{cohen2023calibrating} and an image classification task \cite{bertinetto2018meta}.

\subsection{Benchmarks}
We compare E-ICL+FCP with baselines differing in: (\emph{i}) learning paradigm, (\emph{ii}) calibration strategy, and (\emph{iii}) design criterion:    \textbf{(\emph{i}) Learning paradigm:} Besides ICL, we consider joint learning (JL), which trains a single model across all tasks, and meta-learning via MAML \cite{finn2017model}, which adapts task-specific models.   \textbf{(\emph{ii}) Calibration strategy:} Both SCP and FCP are considered. \textbf{(\emph{iii}) Design criterion:} We contrast standard CP-agnostic objectives (based on log-loss) with the CP-aware objective \eqref{eq:loss_combined}.
All schemes designed in \eqref{eq:loss_combined} are referred with the prefix ``E-''. This results in the following schemes: JL+SCP, E-JL+SCP, MAML+SCP, E-MAML+SCP, MAML+FCP, E-MAML+FCP, ICL+SCP, E-ICL+SCP, and ICL+FCP \cite{huang2025predictions}.

\subsection{Symbol Demodulation}
\noindent\textbf{Task description:} We start with a synthetic example of practical engineering relevance, namely symbol demodulation in wireless communication \cite{cohen2023calibrating}.  
The transmitted symbol $y$ is from the QPSK constellation $\mathcal{Y}=\{-1-j,-1+j,1+j,1-j\}$, where $j$ is the complex unit and  $|\mathcal{Y}|=K=4$.  
The received sample is $x = e^{j\phi} f(y) + v$, where $v\sim \mathcal{CN}(0,\gamma^{-1})$ and $\gamma$ is the signal-to-noise ratio (SNR), where the function $f(\cdot)$ models task-specific deterministic receiver's impairments. 
Specifically, following \cite{tandur2007joint}, this is modeled as  $f(y) = \bar{y}_{\mathrm{I}} + j\bar{y}_{\mathrm{Q}}$, where
\begin{equation}
\begin{bmatrix}
\bar{y}_{\mathrm{I}} \\ \bar{y}_{\mathrm{Q}}
\end{bmatrix}
=
\begin{bmatrix}
1 + \epsilon & 0 \\ 0 & 1 - \epsilon
\end{bmatrix}
\begin{bmatrix}
\cos\delta & -\sin\delta \\ -\sin\delta & \cos\delta
\end{bmatrix}
\begin{bmatrix}
y_{\text{I}} \\ y_{\text{Q}}
\end{bmatrix},
\end{equation}
with $(y_{\text{I}},y_{\text{Q}})$ being the real and imaginary parts of $y$.  
The task-specific parameters are  $(\phi,\epsilon,\delta,\gamma)$, where $\phi\sim\text{Uniform}(0,2\pi)$, $\epsilon\sim\text{Uniform}(0,0.3)$, $\delta\sim\text{Uniform}(0,\pi/6)$, and $10\log_{10}(\gamma)\sim\text{Uniform}(0,10)$.  
\begin{table*}[h]
\centering
\caption{Computational Complexity of different schemes (for $r$ test inputs)}
\label{tab:complexity}
\linespread{1.1}\selectfont
\renewcommand{\arraystretch}{0.9}
\setlength{\tabcolsep}{3pt}
% \footnotesize % 缩小字体
\vspace{-2pt}
\begin{tabular}{c|cc|cc|cc|cc|cc} 
\toprule
 & \makecell{JL\\+SCP} & \makecell{E-JL\\+SCP} 
 & \makecell{MAML\\+SCP} & \makecell{E-MAML\\+SCP} 
 & \makecell{MAML\\+FCP} & \makecell{E-MAML\\+FCP} 
 & \makecell{ICL\\+SCP} & \makecell{E-ICL\\+SCP} 
 & \makecell{ICL\\+FCP} & \makecell{\bf E-ICL\\\bf +FCP} \\ 
\midrule
\# model training      
  & \multicolumn{2}{c|}{/} 
  & \multicolumn{2}{c|}{1} 
  & \multicolumn{2}{c|}{$|\mathcal{Y}|\cdot r$} 
  & \multicolumn{2}{c|}{/} 
  & \multicolumn{2}{c}{/} \\
\# prediction evals     
  & \multicolumn{2}{c|}{$n+r$} 
  & \multicolumn{2}{c|}{$m+r$} 
  & \multicolumn{2}{c|}{$|\mathcal{Y}|\cdot r\cdot (n+1)$} 
  & \multicolumn{2}{c|}{$n+r$} 
  & \multicolumn{2}{c}{$|\mathcal{Y}|\cdot 2\cdot r \cdot(n+1)$} \\
running time (ms, evaluated at $r=1$)
  &0.834
  &0.833
  &15.519
  &15.589
  &59.172
  &60.336
  &2.792
  &2.737
  &2.882
  &2.919\\
\bottomrule
\end{tabular}
\vspace{-10pt}
\end{table*}

\noindent\textbf{Network architectures and training:} 
For ICL-based schemes, we adopt a six-layer Transformer, with an input dimension of 16, two attention heads, and a feed-forward network dimension of 1024, with projectors $h_1(\cdot)$ and $h_2(\cdot)$ implemented as linear layers. 
For the remaining baselines, we employ a four-layer feedforward neural network.

The dataset $\mathcal{D}$ consists of $n = 19$ examples. For SCP, these are  split  SCP into a training set of size $l=10$ and a calibration set of size $m=9$. 
Training and validation leverage 256 tasks each. Each  task is characterized by  50 i.i.d. realizations of data $(\mathcal{D},\mathbf{x}_{n+1})$. 
Finally testing uses 512 tasks with 50 i.i.d. realizations of data $\mathcal{D}$, and 10 additional i.i.d. input $\mathbf{x}_{n+1}$ samples. 
All schemes are trained using an initial learning rate of 0.0002 with a CosineAnnealingLR scheduler (50 iterations, minimum 0.00002) \cite{pytorch_cosineannealinglr_docs} and are trained for 200 epochs.

\noindent\textbf{Results:}
\begin{figure}
    \centering
    \includegraphics[width=1.0\linewidth]{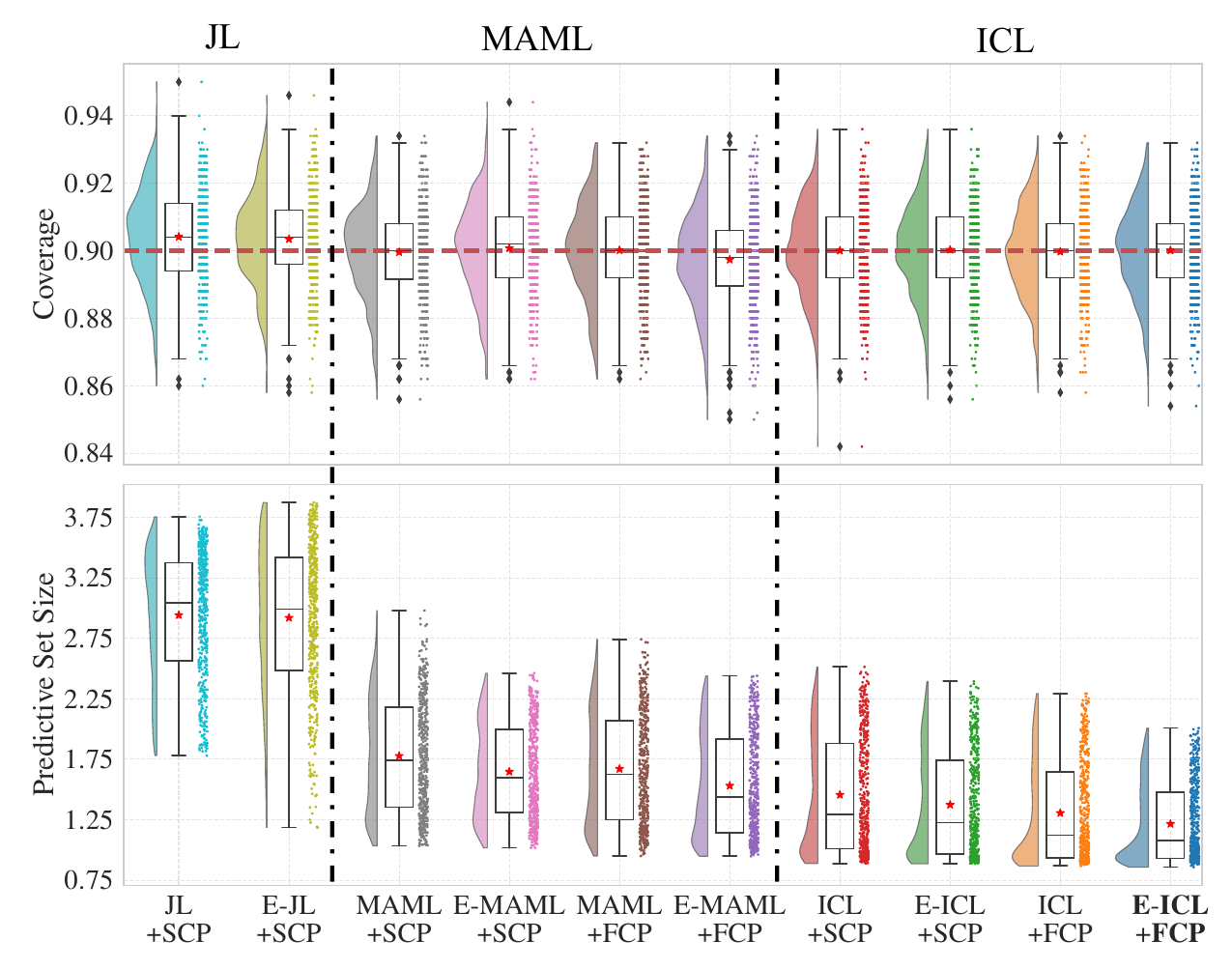}
    \caption{Coverage probability and predictive set size achieved by different schemes on the QPSK demodulation tasks ($\textcolor{red}{\bigstar}$ marks the mean, $\alpha=0.1$).}
    \vspace{-15pt}
    \label{fig:fig_optimized_icl_fcp}
\end{figure}
As shown in Fig.~\ref{fig:fig_optimized_icl_fcp}, all schemes achieve coverage rates around the prescribed level $1-\alpha = 0.9$. 
FCP-based schemes yield smaller predictive sets than their SCP-based counterparts, while schemes optimized with CP-aware loss also produce smaller predictive sets compared to those trained with cross-entropy loss. 
The proposed E-ICL+FCP method achieves the smallest predictive sets, with an average size of 1.216, which is a 6.99\% reduction compared to the state-of-the-art ICL+FCP.

\par Table~\ref{tab:complexity} compares the calibration-phase computational complexity by reporting the number of model training steps and prediction evaluations required by different schemes for $r$ test inputs. 
The computational complexity of E-ICL+FCP scales linearly with $|\mathcal{Y}|$ and $r$, and grows from near-linear to quadratic with $n$.  
Its training-free design bypasses the need for computational- and memory-intensive gradient descent iterations via backpropagation during inference, enabling improved hardware utilization.
We also report the average running time per test input ($r=1$), encompassing both training and evaluation, on a computing platform equipped with a single NVIDIA GeForce RTX 4090 GPU and an Intel(R) Xeon(R) Gold 6240 CPU.
The proposed E-ICL+FCP is seen to offer the best performance-complexity trade-off, producing the smallest predictive sets with relatively low running time.

\subsection{CIFAR-FS Image Classification}
\noindent\textbf{Task description:}
The CIFAR-FS dataset \cite{bertinetto2018meta} consists of 100 object classes, each containing 600 RGB images of size $32 \times 32$. 
Each task is a binary classification problem involving images from two randomly selected classes. 
64 classes are used for meta-training, 16 for validation, and 20 for meta-testing, following the standard few-shot classification protocol.

\noindent\textbf{Network architectures and training:}
The network architectures adopted in ICL-based schemes are unified as a six-layer Transformer encoder, with an input dimension of 128, four attention heads, and a feed-forward network dimension of 1024. The projectors $h_1(\cdot)$ and $h_2(\cdot)$ are implemented as a hybrid 2D CNN followed by a linear layer, and a linear layer, respectively.

\par We consider a number of examples $n = 19$ for the dataset $\mathcal{D}$, which is split by SCP into a training set of size $l=10$ and a calibration set of size $m=9$. 
Training uses 256 tasks with 50 i.i.d. realizations of $(\mathcal{D},\mathbf{x}_{n+1})$ each; validation uses 120 tasks of same form; while testing uses 128 tasks with 50 i.i.d. realizations of $\mathcal{D}$, and 10 additional i.i.d. $\mathbf{x}_{n+1}$ samples per realization. For training, we adopt the same schedule as in the previous example. 

\begin{figure}
    \centering
    \includegraphics[width=0.95\linewidth]{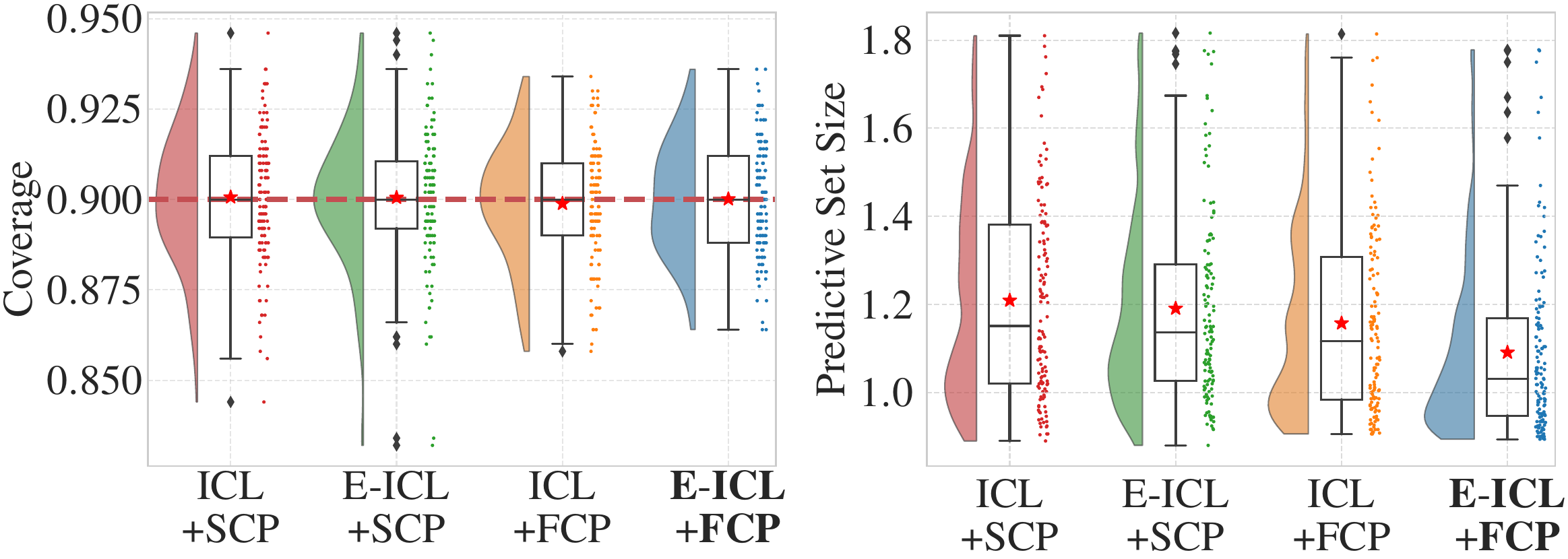}
    \caption{Coverage probability and predictive set size achieved by ICL-based schemes on the image classification tasks ($\textcolor{red}{\bigstar}$ marks the mean, $\alpha=0.1$).}
    \label{fig:coverage_inefficiency_CIFAR_FA}
    \vspace{-15pt}
\end{figure}

\noindent\textbf{Results:}
As in the previous example, the JL-based schemes produce large predictive sets, while the MAML-based schemes are severely limited by their high complexity, hence, we focus on the ICL-based schemes only. 
The results in Fig.~\ref{fig:coverage_inefficiency_CIFAR_FA} are consistent with the previous example, showing that FCP-based schemes achieve smaller predictive sets than SCP-based schemes. 
Moreover, the CP-aware training further reduces the predictive set size achieved by the ICL models. 
In particular, the proposed E-ICL+FCP attains an average predictive set size of 1.091, representing a 5.73\% reduction relative to ICL+FCP.

\section{Conclusions}
In this paper, we have proposed an efficient full conformal prediction framework based on in-context learning. 
The model employs a permutation-invariant Transformer architecture,  and is pre-trained with a CP-aware loss. 
Future work may explore extensions to regression tasks and real-world large-scale applications, including reducing complexity through candidate-label pruning and hierarchical prediction strategies \cite{fisch2020efficient}.

\clearpage
% \balance
\bibliographystyle{IEEEtran}
\bibliography{IEEEabrv,ref}

\end{document}